\documentclass[conference]{IEEEtran}
\IEEEoverridecommandlockouts
\usepackage{times}

\usepackage{soul}
\usepackage{url}
\usepackage[utf8]{inputenc}
\usepackage[T1]{fontenc}
\usepackage[small]{caption}
\usepackage{graphicx}
\usepackage{amsmath}
\usepackage{amsfonts}
\usepackage{bm}
\usepackage{booktabs}
\usepackage{mathtools}

\usepackage{algorithm}
\usepackage{algpseudocode}

\usepackage{color}
\usepackage[normalem]{ulem}
\usepackage{flushend}

\newcommand{\pname}{\texttt{FedGMCC}}
\DeclareMathOperator{\EX}{\mathbb{E}_{\x|y=i}}% expected value
\DeclareMathOperator{\x}{\mathbf{x}}% input dataset
\DeclareMathOperator{\y}{\mathbf{y}}% output dataset
\DeclareMathOperator{\X}{\mathbf{X}}% input distribution
\DeclareMathOperator{\W}{\mathbf{w}}% weights
\DeclareMathOperator{\tta}{\boldsymbol{\theta}}% centroid or center of gravity

\usepackage[
  sorting=none,
  style=numeric-comp,
  backend=biber,
  isbn=false,
  url=true,
  doi=false,
  url=false,
  mincrossrefs=100,
  maxnames=12 %,%%The Nuprl book has 12 authors
]{biblatex}

\AtEveryBibitem{\clearname{editor}}

\usepackage[
  colorlinks=true,
  linkcolor=black,
  citecolor=blue,
  urlcolor=black,
  pdfauthor={},
  pdftitle={},
  pdfsubject={},
  pdfkeywords={},
  bookmarks=false,
]{hyperref}

\begin{document}

\hypersetup{colorlinks=true}

\title{Federated Geometric Monte Carlo Clustering to Counter Non-IID Datasets}

\makeatletter
\newcommand{\linebreakand}{%
  \end{@IEEEauthorhalign}
  \hfill\mbox{}\par
  \mbox{}\hfill\begin{@IEEEauthorhalign}
}
\makeatother

\author{
  \IEEEauthorblockN{
  	Federico Lucchetti\IEEEauthorrefmark{1},
  	Maria Fernandes\IEEEauthorrefmark{2},
  	Lydia Y. Chen\IEEEauthorrefmark{3},
 	Jérémie Decouchant\IEEEauthorrefmark{3}, 
  	Marcus Völp\IEEEauthorrefmark{1}}
  \IEEEauthorblockA{
    \IEEEauthorrefmark{1}SnT - University of Luxembourg, 
    \IEEEauthorrefmark{2}WHG - University of Oxford, 
    \IEEEauthorrefmark{3}Delft University of Technology, \\ 
    \url{federico.lucchetti@uni.lu}, \url{maria.fernandes@well.ox.ac.uk}, \url{j.decouchant@tudelft.nl}, \url{y.chen-10@tudelft.nl},  \url{marcus.voelp@uni.lu}
    }
}

\maketitle

\thispagestyle{plain}
\pagestyle{plain}

\begin{abstract}
Federated learning allows clients to collaboratively train models on datasets that are acquired in different locations and that cannot be exchanged because of their size or regulations. Such collected data is increasingly non-independent and non-identically distributed (non-IID), negatively affecting training accuracy. 
Previous works tried to mitigate the effects of non-IID datasets on training accuracy, focusing mainly on non-IID labels, however practical datasets often also contain non-IID features.
To address both non-IID labels and features, we propose \pname\footnote{
Code and genome dataset available at: \url{https://figshare.com/s/dc2f4280ce012e12f414}
}, a novel framework where a central server aggregates client models that it can cluster together. 
\pname\ clustering relies on a Monte Carlo procedure that samples the output space of client models, infers their position in the weight space on a loss manifold and computes their geometric connection via an affine curve parametrization. \pname\ aggregates connected models along their path connectivity to produce a richer global model, incorporating knowledge of all connected client models.     
\pname\ outperforms FedAvg and FedProx in terms of convergence rates on the EMNIST62 and a genomic sequence classification datasets (by up to +63\%). \pname\ yields  an improved accuracy (+4$\%$) on the genomic dataset with respect to CFL, in high non-IID feature space settings and label incongruency. 
\end{abstract}

\section{Introduction}
\label{sec:introduction}
%========================

Federated learning (FL) frameworks~\cite{mcmahan2016federated,DBLP:conf/ijcai/Yang21} are commonly used when regulations (such as the GDPR\footnote{\url{https://gdpr.eu/}})
or mere data volume prevent data exchanges.
Datasets are distributed across the members (clients) of the federation and the global machine learning optimization problem can approximately be split up into smaller sub-problems that are distributed across independently acting clients. Client solutions are aggregated either by a central server or in a distributed manner. However, this approximation only holds in an idealised settings where client datasets are sampled from the same distribution and data samples form independent events. Deviating from independent and identically distributed datasets (IID) poses a challenge to most FL approaches and results in global models that fail to accurately represent the entire aggregated dataset~\cite{kairouz2019advances,hsieh2020non,zhao2018federated,li2018federated}.
Sources of non-IID-ness can be found in both the label space and the feature space. In the former, for a given feature, different labels might be attributed due to regional differences (e.g., different sentiments in fashion). This is also termed concept shift and leads ultimately to clients with incongruent labelling participating to the training of a common FL model. In addition, classes may be imbalanced due to one class of labels being over-represented with respect to others. This paper focuses on the latter, i.e., \emph{feature space IID violations}. They appear for example when a particular feature is over-represented in a member's dataset compared to other datasets. A biobank might sample genes that bias towards a particular local population~\cite{van2017population} with the consequence of possibly overlooking regional differences, such as the well known preponderance of a specific gene mutation coding for sickle cell disease in the Sub-Saharan Africa population~\cite{williams2016sickle}. For a given label, features might also vary across datasets, such as different handwriting styles for the same alphanumeric character (e.g., 7 with and without bar). 
Non-IID-ness may lead to feature and label
skew~\cite{kairouz2019advances} and accuracy
degradation~\cite{hsieh2020non,li2018federated,li2021federated} by
leaving the choice between starting from
different initialization weights, causing models
to diverge, or accepting diversity suppression when starting from the
same weights~\cite{zhao2018federated}.

Besides statistical heterogeneity, FL algorithms need to cope with
system heterogeneity, leading to asynchronous model updates or incomplete
local training.
The first proposed FL attempt, FedAvg~\cite{mcmahan2016federated}, aggregates all client models by averaging their weights, leading to the above inaccuracies in the presence of non-IID-ness.
Variations~\cite{li2018federated,li2021fedbn} of FedAvg have been
proposed, with satisfactory results when the non-IID-ness is less
pronounced and exclusively in the label space (see
Sec.~\ref{sec:related_work}).  
Federated Clustering~\cite{ghosh2020efficient,briggs2020federated,sattler2020clustered,kopparapu2020fedfmc} 
tackles this problem by assigning clients
to separate clusters and hence limiting the deleterious transfers of negative knowledge between member models that have been trained on distinct datasets. These approaches result in the construction of trained models with reasonable accuracy despite IID violations,
however, at the cost of significant communication and/or computational overheads.

In this paper, focusing on non-IID-ness in the feature space, we propose a novel
Monte Carlo Clustering approach, called \pname, to counter non-IID-ness without
compromising final model accuracy and while maintaining low communication costs and data privacy. We show that in order to train a global FL model in a non-IID setting, the objective function to be minimized has to be conceptually split up into two components: the first encompasses the IID component for which the usual FedAvg solution holds; the second captures the non-IID contribution, which we show can be solved by introducing \emph{interaction} between client models. We derive this interaction by leveraging an observation
about the geometry of training loss manifolds
~\cite{garipov2018loss,freeman2016topology,mirzadeh2020linear,choromanska2015loss}, namely that seemingly different,
stationary solutions to the training loss minimization problem, obtained at the individual member sites, are often connected via simple
parametric curves where the training loss is approximately flat. The existence of these curves reveals the pair-wise interaction between models needed to solve the non-IID problem, which leads us to establish a criterion for clustering seemingly different solutions and suppressing negative knowledge transfer between non-connected ones. More importantly, we show that drawing from~\cite{izmailov2018averaging}, averaging models along the curve parametrization, can produce global models with enhanced accuracy and generalization. As a summary, we make the following contributions.
\begin{enumerate}
    \item We separate the FL objective in an IID and a non-IID components and, based on the geometry of curved training loss manifolds, put forward an Ansatz solution that simultaneously minimizes both components.
    \item We demonstrate how 
    to construct this solution using a Monte Carlo sampling of the received client model output spaces, and present novel model weight clustering and aggregation rules.
	\item We evaluate \pname\ using the EMNIST62 dataset and a genomic dataset with different non-IID-ness. \pname\ always outperforms FedAvg~\cite{mcmahan2016federated} and FedProx~\cite{li2018federated}. It outperforms CFL~\cite{sattler2020clustered} in case of high non-IID-ness. 
\end{enumerate}

\section{Related Work}
\label{sec:related_work}

Among the myriad of published FL algorithms, most of them rely on clients to upload model weight updates onto a central server that proceeds to aggregate them and to redistribute the result back to clients. The main difference often relies in the way the aggregation rule is applied. 
FedAvg~\cite{mcmahan2016federated} was first to allow clients to train a global model
locally, instead of transferring data. FedAvg returns to a central server
only for aggregating all local models by averaging their weights.
FedProx~\cite{li2018federated} inhibits local updates, by adding a
proximal term to the loss function, which restrains divergence between
local and global model weights.
FedBn~\cite{li2021fedbn} achieves accurate results on non-IID datasets by batch normalizing the local neural networks' input layer before aggregation.
FedFV~\cite{wang2021federated} detects inconsistent gradient updates and corrects them before the aggregation step. 
Our approach (\pname) aggregates models by averaging weights along their geometric connection.

Federated clustering groups certain client models into separate clusters to prevent negative knowledge transfer such as the iterative federated clustering
algorithm (IFCA)~\cite{ghosh2020efficient} which solves an incongruency
that occurs when non-IID datasets exhibit 
concept shifts  by
training multiple models on local data and returning the one with the lowest 
aggregation loss.
However, these works primarily focus on label-space non-IID-ness, leading to inaccurate results in the presence of feature-space non-IID-ness. Moreover, 
training
multiple models induce high communication and computation costs.

Merging local models with approximately the same weights (according to
cosine similarity, L1 or L2 norm) reduces the number of models in the
ensemble~\cite{briggs2020federated,sattler2020clustered}. For example,
clustered FL (CFL)~\cite{sattler2020clustered} merges models if their weights (or gradients) compare well enough based on a specified metric. Again CFL focuses on label non-IID-ness. In contrast, \pname\ explicitly considers feature-space non-IID-ness. 
Among the aforementioned FL approaches, CFL is the most reminiscent to our approach, 
with two differences. First in CFL, the central server applies clustering after multiple FedAvg iterations whereas \pname\ clusters after each client-server communicating round. Second, CFL's clustering rule relies on measuring whether the gradients of individual client-weight models are coherent (parallel) via the cosine similarity measure. Our approach verifies that a parallel transport of one client model weights to the other is feasible. Hence, \pname\ extends CFL's gradient coherency measure to include intermediary models along the affine connection. 

\section{Problem Description}
\label{sec:problem_formulation}
%========================

We consider $K$ clients that aim to 
locally minimize a local objective $\mathcal{L}_k{=}\mathcal{L}(\X_k, \W)$.   Clients send their respective solutions (client model weights) $\W_k^{*} {=} \arg \min_{\W} \mathcal{L}_k$ to a central server to 
approximate with aggregate $\W_{t}^{f}$ a global solution for the loss
minimization problem $\W^{*} {=} \arg \min_{\W} \mathcal{L}(\x, \W)$,
which 
maps the $C$ classes of the compact input space $\mathcal{X}$
into the label space $\mathcal{Y} = [\mathbf{C}]$, where $[\mathbf{C}]={1,...,C}$. 
We consider the neural network as a function
$f(\W):\mathcal{X} \mapsto \mathcal{S}$, parametrized by weights $\W$,
that maps $\X$ to the probability simplex 
$\mathcal{S}{=}\left\{ \mathbf{z} | \sum_{i=1}^{C} z_i=1, z_i \geq 0, \forall i \in
[\mathbf{C}] \right\}$. We write $f_i$ for the probability of the $i$-th class and
define the population loss $\mathcal{L}(\X_k, \W_k)$ at federation
member $k$ as the cross-entropy loss $\mathcal{L}(\X_k,\W) {=} \frac{1}{N} \sum_{x_i \in \X} \sum_{j=1}^{C} p(y{=}j) \EX \left[  \log f_j(x_i,\W_k)  \right]$.
In an IID setting, the optimization problem minimizing $\mathcal{L}_{IID}$ can be expressed as $K$
sub-problems that train local models via stochastic gradient descent (SGD), which are
then communicated to a server for aggregation after each communication
round (indexed by $t$):
\begin{equation}
	\label{eq:fl_optimization_iid}
	\mathcal{L}_{IID} = \sum_{k=1}^{K} \frac{n_k}{n} \mathcal{L}(\X_k,\W) = \mathbb{E}_{\X_k}(\mathcal{L}(\X_k,\W))
\end{equation}
FedAvg~\cite{mcmahan2016federated} makes use
of the weighted averaging aggregation function $\W_{t}^{f}
= \sum_{k}^{K} \frac{n_k}{n} \W_{t,k}^{*}$, where $\frac{n_k}{n}$ is
the fraction of the $k$-th member's local dataset with repect to the
overall dataset size. 
Unfortunately, in a non-IID setting, this aggregation is not a valid approximation.

We follow~\cite{zhao2018federated} in quantifying non-IID-ness with the help 
of the Earth Mover Distance (EMD)~\cite{rubner1998metric} and write $d(\mathcal{D}_1,...,\mathcal{D}_K)$
for the EMD of $K$ distributions $\mathcal{D}_1,...,\mathcal{D}_K$ obtained by averaging the set of pairwise EMDs (cf. Apx.~\ref{apx:emd}).

Weighted averaging fails as an aggregation function under non-IID datasets~\cite{zhao2018federated} because: 
(1) if weights in client models are all initialized
identically, the label-space non-IID-ness between client dataset distributions will be the
dominant factor and weights will diverge; and (2) if weights are
initialized differently, client models risk converging to different
solutions \footnote{Without a formal proof,a third contribution term  to weight divergence could be added in terms of feature-space non-iid}.
The steepness of local loss function pockets in relation to the local
flatness of the loss manifold exacerbates the divergence after
aggregation and hence leads to sub-optimal global model
performance~\cite{keskar2016large}, which we address with 
Geometric Monte Carlo Clustering.

\section{Geometric Monte Carlo Clustering}
\label{sec:gmcc}

In a non-IID setting, individual client models are set to converge during training towards distinct local minima, i.e., different coordinates in weight space on the training loss manifold (see Fig.~\ref{fig:mcclustering}). In particular, when the non-IID-ness lies exclusively in the feature space between two client datasets $\{\X_1,\y_1\}$ and $\{\X_2,\y_2\}$, we hypothesize the existence of a continuous transformation $\mathcal{T}$ (e.g. pixel rotation, color inversion, shape distortion, etc.) 
that maps, on average, one subset of features $\X_1' {\in} \X_1$ detained by one client to a subset of features $\X_2' {\in} \X_2$ of a different client $\mathcal{T}{:} \X_1 {\rightarrow} \X_2$. As a consequence, we suppose that the model weights of one client are trained to encode a set of transformed features with respect to another client.  Hence we hypothesize the existence of a continuous transformation $\Gamma$ that maps one subset of weights to another one $\Gamma: \W'_1 \subseteq \W_1 \rightarrow \W'_1 \subseteq \W_1$ and that can alternatively be modeled by a continuous affine connection via a curve parametrization $\gamma_\theta(u)$. This curve connects $\W_1$ and $\W_2$ (with $\gamma_\theta(0){=}\W_1$ and $\gamma_\theta(1){=}\W_2$) ideally on a loss surface where the loss value does not vary. That is, two neural nets parametrized by $\W_1$ and $\W_2$ agree on the output space given the same input. Along this curve of invariant loss reside a family of intermediary model weights that can be aggregated to produce a richer global model, incorporating the knowledge of both client models. In contrast, under label-space IID violations, particularly concept shifts, we expect no continuous curve to be found (e.g. between $\W_3$ and $\W_4$ on Fig.~\ref{fig:mcclustering}) since they disagree on the classification of a same input feature. In this case, a well constructed clustering rule should separate both models to avoid negative knowledge transfer. This curve finding is at the heart of our novel Federated Geometric Monte Carlo Clustering algorithm (\pname).

\begin{figure}[h!]
	\centering
	\includegraphics[width=0.8\linewidth]{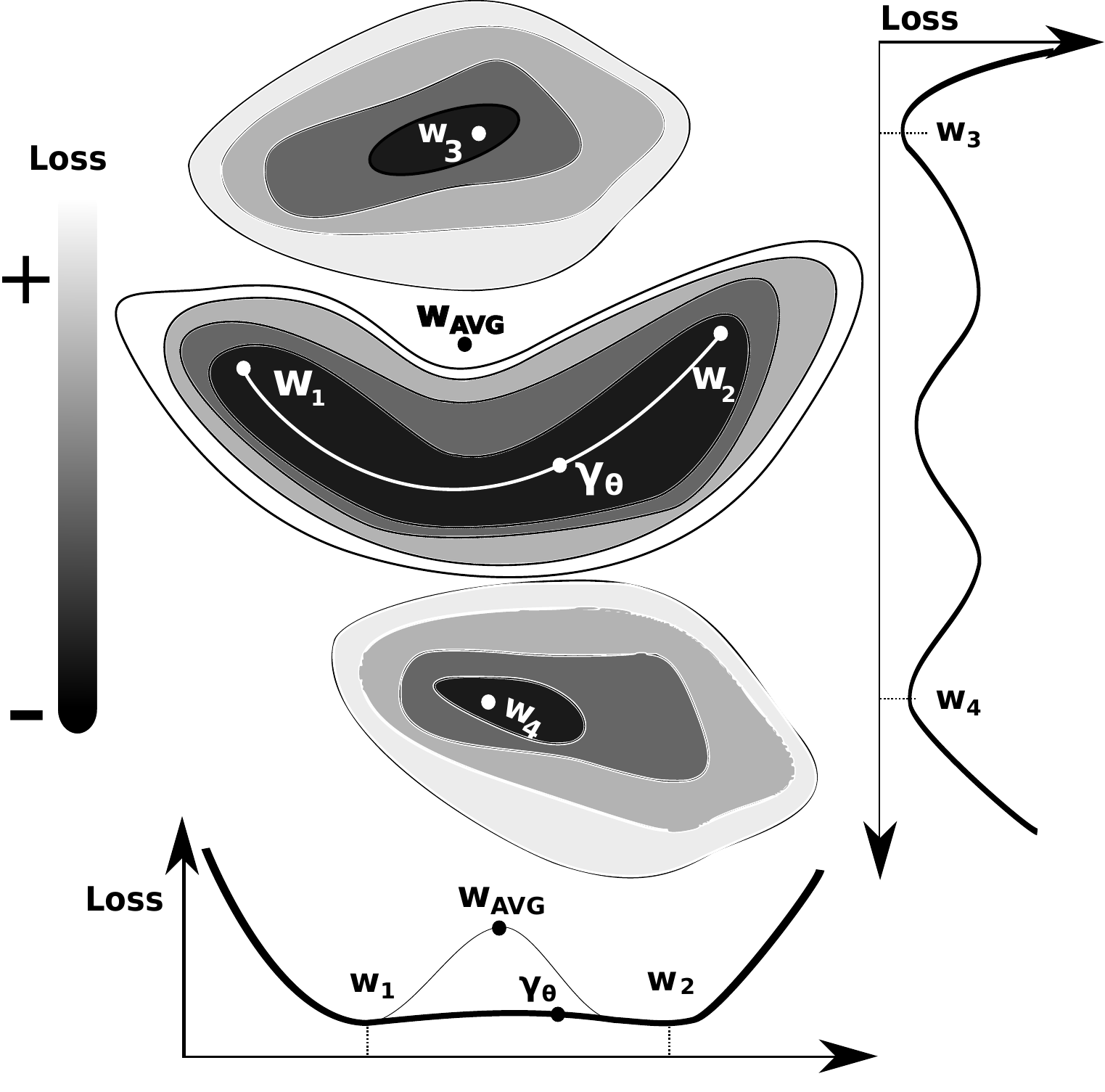}
	\caption{Schematatization of the Geometric Monte Carlo Clustering applied on 4 trained clienz models where 1 and 2 have a feature space based IID, 3 and 4 have a label space concept shift based non-IID. Curved line connects $\W_1$ to $\W_2$ where the loss remains minimally low. No curve is found between $\W_3$ to $\W_4$ where the loss crossed the $\epsilon$ budget. Note that because of the curved loss surface, the standard FedAvg aggregation $\W_{AVG} = 0.5 \cdot (\W_1 + \W_2)$ would lead to a suboptimal model (high loss).}
	\label{fig:mcclustering}
\end{figure}

\subsection{Prerequisites}

Key to \pname\ is the treatment of non-IID datasets as perturbations of probability distributions $\mathcal{D}_k$  from the ideal IID baseline $\mathcal{D}$, where  $\mathcal{D}_k \mapsto \mathcal{D} {+} \delta_k$ and $d(\mathcal{D}_k,\mathcal{D}) {\neq} 0$.
Datasets that originally had overlapping feature and label representations have, after perturbation, a component that contributes to their non-IID-ness. Let us denote as $\{\X_k^U,\y_k^U\}$ the labeled dataset drawn from the distribution $\mathcal{D}_k^U$ such that $\forall k {\neq} l, d(\mathcal{D}_k^U,\mathcal{D}_l^U){=}0$ and as $\{\X_k^\delta,\y_k^\delta\}$ the datasets drawn from $\mathcal{D}_k^\delta$ such that $\forall k {\neq} l, d(\mathcal{D}_k^\delta,\mathcal{D}_l^\delta)\neq 0$, where  $\{\X_k^U,\y_k^U\} \cup \{\X_k^\delta,\y_k^\delta\} =  \{\X_k',\y_k'\}$. The training objective is:
\begin{equation}
	\label{eq:noniid_loss}
	\begin{array}{l}
	\sum_{k}^{K} \mathcal{L}(\tilde{\X}_k,\W) \\\qquad
	\approx \sum_{k}^{K} \mathcal{L}(\X_k^U,\W) +\sum_{k <l}^{K} \mathcal{L}(\X_k^\delta \cup \X_l^\delta,\gamma_{k,l}) \\
		\qquad= \mathcal{L}_{IID} + \mathcal{L}_{INT}
	\end{array}
\end{equation}
$\mathcal{L}_{IID}$ is the usual IID loss function
(see Eq.~\ref{eq:fl_optimization_iid}).
The \emph{pairwise interaction loss} $\mathcal{L}_{INT}$ captures the training loss on a pair of non-IID distributions.
We use the solutions $\W_k$ to the pure IID objective to approximate the solution to $\min_{\W} \mathcal{L}_{INT}$.

\subsection{Curve Finding} We intuit the solution to the minimization of Eq.~\ref{eq:noniid_loss} to be of the form $\W{=}(1-u)\W^f + u \tta_{k,l}$ 
where $\tta_{k,l}$ introduces a pairwise \emph{interaction} between two client model weights and 
balances its contribution to the total solution with a coupling constant $u {\in} [0,1]$. As we have already hinted in the previous section on the possibility of connecting distinct model weights via a transformation, we propose to model this interaction as an affine connection 
between weights $\W_k$ and $\W_l$ by a smooth curve $ \gamma_{\tta}(u) {:} [0,1] {\mapsto} \mathcal{M} $ on the loss function manifold
 $\mathcal{M}$ of dimension $|net|$ equal to the number of parameters of the neural net and parametrized by $\tta {\in} \mathbb{R}^{|net|}$. 
 We need to find the transport parameter $\tta$ that leaves the gradient of the second term in Eq.~\ref{eq:noniid_loss} parallel
for every $u {\in} [0,1]$. This amounts to solving the geodesic equation $\partial_u\nabla_{\tta} \mathcal{L}(\X_k^\delta \cup \X_l^\delta,\gamma_{\tta}(u)){=}0$ 
with boundary conditions  $\gamma_{\tta}(0){=}\W_k$ and $\gamma_{\tta}(1){=}\W_l$. Ideally, the central server, which detains the individual client model weights, should execute the curve finding procedure  after every local update. However, since the central server
does not have access to the datasets in order to explore the loss manifold, we sample a surrogate version of the latter by generating a Monte 
Carlo type input dataset drawn from a uniform distribution $\X_{MC} \sim U(0,1)$ to reconstruct the loss from the label-space output $f(\X_{MC},\W_k)$, using
mean-squared error (MSE) as loss function, since we merely compare raw outputs.
The curve is then derived by perturbing the parameter $\tta$  in the direction where the loss 
$\mathcal{L}_{MSE}(\W_l,\gamma_{\tta}(u))$ does not vary. We rely on a polygonal chain as Ansatz, as it has been proven to lead to optimal curve finding results \cite{garipov2018loss}.
\begin{equation}
	\label{eq:chain}
	\gamma_{\tta}(u) =
	\begin{cases}
		2(u \tta+(0.5-u)\W_k) & u \in [0,0.5[ \\
		2((u-0.5)\W_l)+(1-u)\tta & u \in [0.5,1] \\
	\end{cases}
\end{equation}
Following ~\cite{garipov2018loss}, we simplify the loss to be minimized to the expectation of $\mathcal{L}_{MSE}(\W_l,\gamma_{\tta}(u))$ with respect to uniform distribution on the curve on $u \in [0,1]$, making the loss computationally tractable.
\begin{equation}
\begin{array}{l}
	\nabla_{\tta} \int_{0}^{1} \mathcal{L}_{MSE}(\W_l,\gamma_{\tta}(u))  du = \\
	\qquad\qquad \nabla_{\tta} \mathbb{E}_{u \sim U(0,1)}  \mathcal{L}_{MSE}(\W_l,\gamma_{\tta}(u))
	\end{array}
\end{equation}

\subsection{Model Weights Clustering and Aggregation}

Garipov et. al \cite{garipov2018loss} showed that two models initialized differently and trained on the same dataset can be connected via a simple curve. Two models trained independently on distributed datasets with only partially overlapping features are also expected to converge to different stationary solutions. Nevertheless, we expect these seemingly different solutions to be interlinked via an affine connection in weight space i.e a smooth curve can be constructed that along a approximately flat loss manifold. Hence, seeminly different solution form a family of equivalent solutions to the loss minimization problem which leads us to put forward the following clustering rule.\\

\noindent \textbf{Proposition 1.1}
\textit{$\X_{MC}  \sim U(0,1)$, two models with weights $\W_k$ and $\W_l$  belong to the same cluster $\mathcal{S}$ if there exists a $\tta \in \mathbb{R}^{|net|}$ that parametrizes the curve  $\gamma_{\tta} : u \in [0,1] \mapsto \mathbb{R}^{|net|}$ and $\epsilon \geq 0$ such that}
\begin{equation}
	\label{eq:cluster_condition}
	 \partial_u \mathbb{E}_{x \in \X_{MC}} \Arrowvert f(x,\W_l) - f(x,\gamma_{\tta}(u))\Arrowvert^2 < \epsilon
\end{equation}
% \textline
Because $\gamma_{\tta}(u)$ generates a family of intermediary model weights, all approximate solutions to the optimization problem (see Eq.~\ref{eq:noniid_loss}), their sum must also be a solution.
\cite{izmailov2018averaging} showed that averaging intermediary model weights on the constructed curve $\gamma_{\tta}(t)$ along $t$ can lead to an aggregated model with improved generalization. \\

\noindent \textbf{Proposition 1.2}
\textit{Model weights $\W_j$ belonging to the same cluster $\mathcal{S}_j$ are aggregated with}
\begin{equation}
	\W_j = \sum_{k,l \in \mathcal{S}_j} \mathbb{E}_{u \in [0,1]}\gamma_{\tta_{k,l}}(u)
\end{equation}

\subsection{Algorithm}

\pname\ is described in Alg.~\ref{algo:fedmcc} and schematized in Fig.~\ref{fig:mcclustering}. The server samples the output space of each received client model by generating a Monte Carlo input dataset and tests via Prop. 1.1 whether two models can be grouped together. This procedure can be seen as a disjoint-set query in which model weights that could not be connected to any other model weights are kept as singleton sets (e.g., $\{ \W_1, \W_2 \},\{ \W_3 \}, \{ \W_4 \} $ on Fig.~\ref{fig:mcclustering}). With $M$ disjoint sets, the weights that belong to a same cluster $\mathcal{S}_j$ are aggregated by averaging all pairwise interactions leading to $M$ clustered global solutions to the non-IID problem (see Proposition 1.2). $\W_j$ is then sent to the clients that contributed to it.

\begin{algorithm}[t]
	\caption{\textit{Geometric Monte Carlo Clustering}}
	\label{algo:fedmcc}
	\begin{algorithmic}[1]
		\footnotesize
		%%%%%%%%%%% Parameters %%%%%%%%%%%
		\State \textbf{Parameters:} $\epsilon>0$; $n$: Monte Carlo sample size; $K$: number of client models;  $M$: number of clusters; $\eta$: learning rate
		%%%%%%%%%%% Initialization %%%%%%%%%%%
		\State \textbf{Initialization:} Random input dataset $\X \leftarrow $ Uniform$(0,1,n)$
		%%%%%%%%%%% Model Clustering %%%%%%%%%%%
		\State \textbf{Server: ModelClustering}$(\W_1,...,\W_K)$:
		\State \hspace*{2em} clusters $\mathcal{S} \leftarrow \{\}$
		\State \hspace*{2em} \textbf{for} $k \in 1,...,K$, $l \in k+1,...,K$  \textbf{do}
		\State \hspace*{4em} \textbf{if} $(\{\W_k,\W_l \} \in \mathcal{S}_{j\leq k})$ \textbf{continue}
		\State \hspace*{4em} \textbf{else} 
		\State \hspace*{6em} $\tta_{k,l} \leftarrow$\textbf{Update}$(\W_k,\W_l)$
		\State \hspace*{6em} \textbf{if} $(\tta_{k,l})$ $\mathcal{S}_j \leftarrow \mathcal{S}_j  \{ \W_k,\tta_{k,l} \}$
		\State \hspace*{2em} $\W_1,...,\W_M \leftarrow$ \textbf{Aggregation}$(\mathcal{S})$
		\State \hspace*{2em} \textbf{Distribute} $\W_1,...,\W_M$
		%%%%%%%%%%% Update %%%%%%%%%%%
		\State \textbf{Update}$(\W_1,\W_2)$:
		\State \hspace*{2em} $\tta \leftarrow \W_1+\W_2 $
		\State \hspace*{2em} \textbf{for} $ u \in U(0,1)$, $ x \in \X_{MC}$ \textbf{do}
				\State \hspace*{4em} $\mathcal{L}(x,\W_2,\gamma_{\tta}(u))\leftarrow\Arrowvert f(x,\W_2)-f(x,\gamma_{\tta}(u))\Arrowvert^2$
				\State \hspace*{4em} $\tta_{u+1} \leftarrow \tta_u - \eta \nabla_{\tta} \mathcal{L}(x,\W_2,\gamma_{\tta}(u))$
		
		\State \hspace*{2em} \textbf{if} $(\max_u \partial_u \mathbb{E}_{x \in \X_{MC}} \mathcal{L}  \leq \epsilon)$ \textbf{return} $\tta$

		%%%%%%%%%%% Aggregation %%%%%%%%%%%
		\State \textbf{Aggregation}$(\mathcal{S}_{1,..,M})$:
		\State \hspace*{2em} \textbf{for} $ S_j \in S$ \textbf{do} $\{\W_j \leftarrow \sum_{k,l \in \mathcal{S}_j} \mathbb{E}_{u \in U(0,1) } \gamma_{\tta_{k,l}}(u) \}$
		\State \hspace*{2em} \textbf{return} $\W_1,...,\W_M$
		
		%%%%%%%%%%% Distribute %%%%%%%%%%%
		\State \textbf{Distribute}$(\W_1,...,\W_M)$:
		\State \hspace*{2em} \textbf{for} $S_j \in S$, $k \in 1,...,K$ \textbf{do}
		\State \hspace*{4em} \textbf{if} $(k \in S_j)$
		    send  $\W_j$ to client $k$  
		
	\end{algorithmic}
\end{algorithm}

\section{Evaluation}
\label{sec:evaluation}
%======================
To evaluate our approach, we show that real-life data is in fact
non-IID in the feature space and compare the accuracy of our approach --- Federated
Geometric Monte-Carlo Clustering (\pname) --- against state-of-the art
federated learning approaches (FedAVG~\cite{mcmahan2016federated},
FedProx~\cite{li2018federated}, CFL~\cite{sattler2020clustered} and
standard SGD). We leverage the well known image dataset EMNIST62~\cite{cohen_afshar_tapson_schaik_2017} and two sequential genomic datasets (SENSG-A and SENSG-R) we created (see Appx.~\ref{app:datasets}). EMNIST62 contains 814255 handwritten characters, labeled as 62 unbalanced classes in 28x28 pixel format. The genomic dataset contains reads (i.e., sequences of the bases A, T, C, G) of size 150 and every position is labelled according to whether or not it is part of a genomic variation. Such labelling is important, for example, to filter out and protect private information in the genomic information processing pipeline~\cite{decouchant2018filter}. We introduce a concept shift in SENSG-R by randomly flipping the label of the most recurrent genomic features. EMNSIT62 and SENSG-A were distributed among $K$ clients following the procedure detailed in  Appx.~\ref{app:datasets}. This partitioning introduces non-IID-ness in the feature space but also in the label space because certain classes might be over-represented in some client datasets compared to others. The genomic dataset SENSG-R is naturally partitioned using the populations of individuals (Asian, European, African). In addition, we partition both sets artificially into 10 client datasets to consider a wider range of non-IID-ness.
Models were trained on an AMD Ryzen7 3700x system with 8 3.6 GHz cores,  a NVIDIA Geforce RTX 3090 GPU with 10496 CUDA cores and 24 GB of GDDR6X memory. We used the binary-cross entropy loss
function for the training of all classifiers and Tensorflow 2.6~\cite{abadi2016tensorflow}. 

\subsection{Network Architecture and Baselines}

We construct two neural network architectures for the 
benchmarks: for EMNIST62, we use 2 stacked CNN layers,
activated by ReLu, and followed by a fully connected neural
network.  The genomic datasets are classified with a network comprised of 2 stacked bidirectional LSTMs that feed into a densely connected
neural network. We compare our approach \pname\ against three federated learning
methods: FedAvg~\cite{mcmahan2016federated}, FedProx~\cite{li2018federated} and CFL~\cite{sattler2020clustered}. We assume a federation of $K = 10$
members for all experiments, each solving its local optimization
problem using SGD. FedAvg performs weighted
average aggregation in the central server after each local training
round, which is performed on the member datasets. FedProx~\cite{li2018federated}
corrects FedAvg's loss function by the proximal term $\mu \| \W
- \W_t\|^2$, where $\mu$ is a positive constant, $\W$ the weights of
the most recent global model and $\W_t$ the weights of the local model
at training step $t$. We implement CFL
by following the procedure laid out in~\cite{sattler2020clustered}.
With FedAvg the central server receives, aggregates and distributes client updates until the average of received client gradients decreases below $\epsilon_1 = 0.2$. \pname\ applies the clustering, aggregation and distribution scheme laid out in Alg.~\ref{algo:fedmcc}.
In the first round, the value of $\epsilon$ was set to the median value of the $\frac{1}{2}K(K-1)$ training losses associated with the curve finding procedure. This value is subject to a 5 percentile increase if the number of clusters was higher than 1 and a 5 percentile decreases otherwise. The learning rate $\eta$ in the curve finding procedure was set to 0.1. In addition, we train two central models ($cSDG_0$ and
$cSDG_1$) on the combined genomic dataset using standard SGD as baselines. Training of two
models is necessary due to the concept shift we introduce in
SENSG-A. As usual, we separate the whole dataset into disjoint
training (80\%) and validation sets (20\%). Hyperparameters have been fine-tuned for every training algorithm to give best possible results in terms of convergence rate. The mini-batch size was 
 set to 64, and learning rates were set to 0.001.
 Local epoch numbers were respectively set to 10 and 5 for the EMNIST62 and SENSG-A datasets.   
 
 \begin{figure}[t]
  \centering
  \includegraphics[width=\columnwidth]{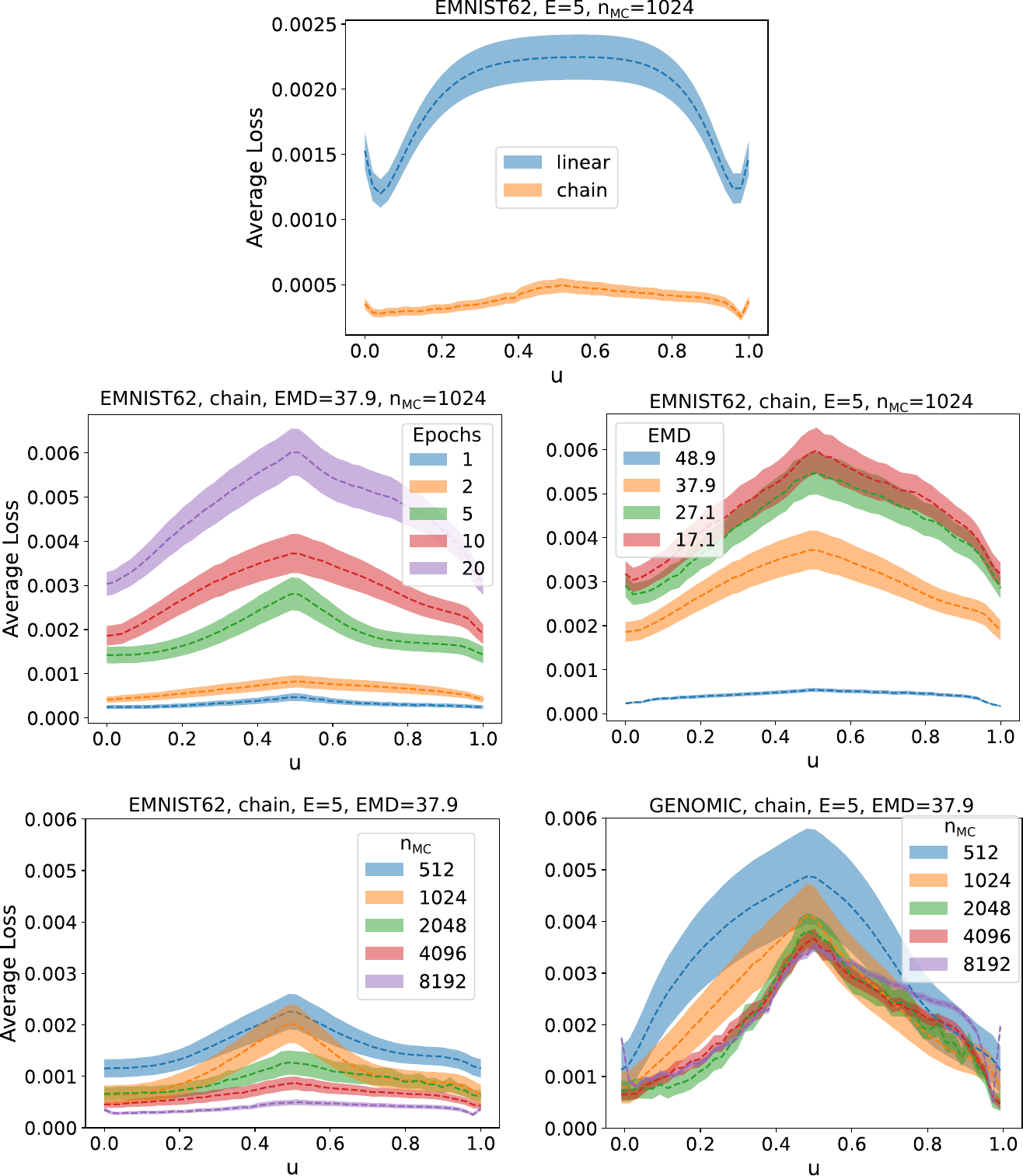}
  \caption{The 10 client averaged train loss as along the curve $\gamma_{\tta}(u)$ connecting two client models as a function of multiple experimental setting (EMD, local epochs, datasets) and curve parameters (type, Monte Carlo sample size $n_{MC}$). Dispersion represents the standard deviation scaled down by a factor of 10. }
  \label{fig:Curve_finding}
\end{figure}
\subsection{Results}

\textbf{Curve Fitting between Local Models:}
Fig.~\ref{fig:Curve_finding} shows the average training loss for our curve-fitting approach for different parameters and setups.
The top graph compares a naive linear curve $\gamma(t) = u \W_1 + (1-u)\W_2$ with a polygonal one-bend chain curve (see~\ref{eq:chain}). Using the latter we were able to find a region of low loss and hence prove the existence of multiple simple connections between pairs of client models, i.e., pairs of client models belonging to the same cluster. Note that by setting $u=0.5$, the linear parametrization reduces to the classical FedAvg aggregated model $\W_1 + \W_2$ situated in a region of higher loss with respect to the chain curve loss $\gamma_{\tta}(u=0.5)$. 
When clients train local models with varying local epoch numbers (middle left) or EMD values (middle right) the average loss is directly affected. When the number of local epochs is decreased or the data distribution setting tends towards the ideal IID setting, the GMCC losses flatten out and the inter-client variability decreases. This is an important finding that could previously only be obtained when computing loss surface manifold with real datasets. Our approach achieves this on a generated dataset using Monte Carlo sampling. More interestingly, increasing the sample size of the GMCC input dataset makes it easier to find connections between client model weights with low training losses. For EMNIST62, where no concept shift was present (bottom left), finding an optimal curve parametrization is only a matter of increasing computational costs (due to increased GMCC sample sizes). On the other hand, for the genomic datasets, because the concept-shift introduced an incongruency in the output space between two models, no amount of GMCC sample size will be able to connect models hence leading to the splitting of client models into separate clusters. We validate our approach, where the server substitutes the client datasets by a surrogate Monte Carlo type generated dataset.

\begin{figure}[t]
	\centering
	\includegraphics[width=.92\linewidth]{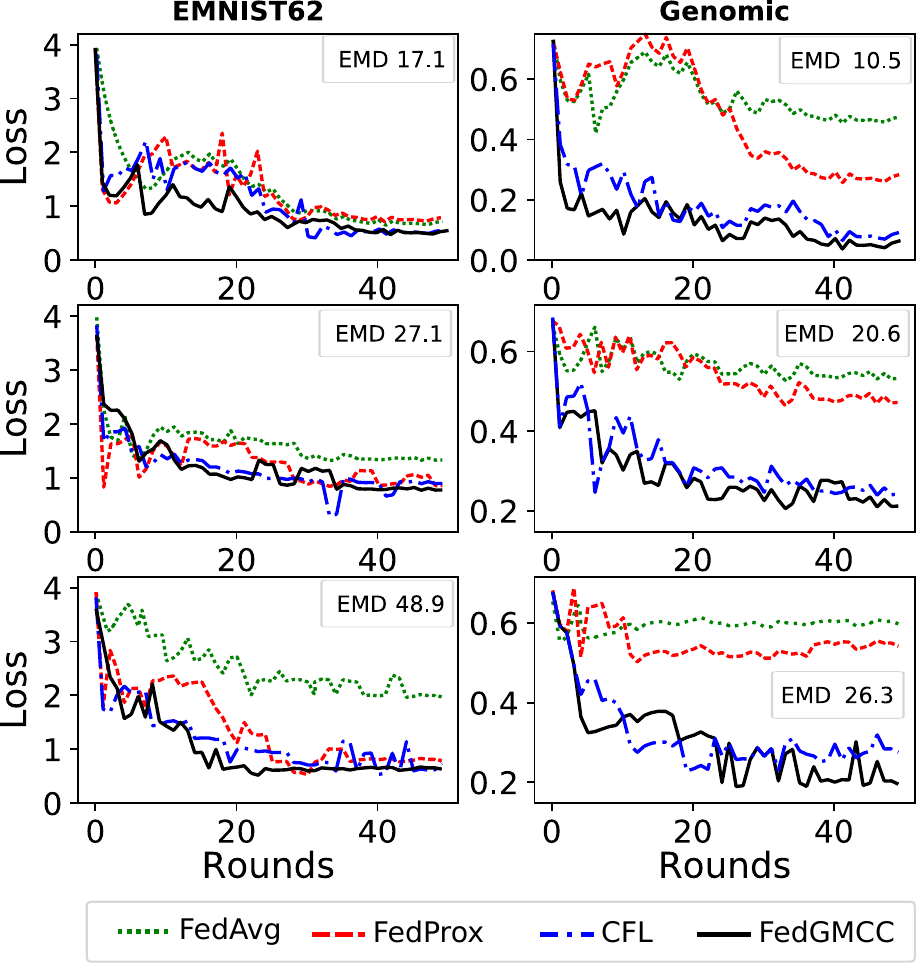}
	\caption{Validation loss for FedAvg, FedProx, CFL and \pname for three different EMD values.}
	\label{fig:Loss_convergence}
\end{figure}

\textbf{Loss:}
Fig.~\ref{fig:Loss_convergence} reports the loss of the FL models for varying EMD values and for the EMNIST62 and genomic datasets. Loss is computed on the validation set after every communication round. For CFL and \pname, where the aggregation can lead to multiple models, the average loss is shown. We highlight the increasing difficulty for FL models to converge when the EMD increases. As expected, FedAvg shows the worst results for both datasets. FedProx is able to deal with high EMD values in the EMNIST62 dataset but does not converge to a lower loss value at high EMD compared to the FL clustering algorithms (CLF and \pname). The latter are able to cluster dissimilar models and avoid negative transfer of knowledge between them, which allows them to obtain the two aggregated final models with optimal loss.

\textbf{Accuracy:}
For the EMNIST62 dataset the FL algorithms achieved a similar performance with low EMD values than the one obtained in the centralized setting (85$\%$) (see Table~\ref{table:testaccuracy}). However, their accuracy degrades quickly when the EMD increases, with values for FedAVG that dropped from 0.81 at EMD 17.1 to 0.43 at EMD 48.9. This drop was less pronounced for FedProx, CFL and \pname. The two latter maintained a 0.79 accuracy at high EMD.
The situation was different for the genomic datasets. The centralized SGD models cSGD$_0$ and cSGD$_1$ achieved a 0.88 accuracy. CFL and \pname\ attained comparable 0.85 accuracy, while FedProx and FedAvg respectively obtained 0.71 and 0.75 at low EMD. As expected, FedAvg's performance dramatically degraded at high EMD to 0.18. The drop in accuracy was less severe for FedProx and stayed around 0.79. \pname\ led to two global models, each of which maintained an accuracy of 0.85 at EMD 17.1 and 0.84 at higher EMD values. This high performance was also achieved by CFL but \pname\ yielded a better accuracy at high EMD. This is not surprising because of CFL's and \pname's aggregation rules, which enable them to create multiple personalized models to accommodate a certain degree of  IID-ness in particular the incongruency due to the concept shift injected in the genomic dataset.

\begin{table}
\caption{FL algorithms accuracies in different EMD settings. Pair of values for CFL and FedGMCC indicate accuracies associated to the final weight clusters.}
\label{table:testaccuracy}
\footnotesize
	\begin{tabular}{|c|c|c|c|c|}
		
		\multicolumn{5}{l}{EMNIST62} \\
		\hline
		EMD & FedAvg & FedProx & CFL & FedGMCC  \\
		\hline
		17.1 & 0.81 & 0.83 & 0.83 & 0.83        \\
		\hline
		27.1 & 0.76 & 0.83 & 0.83 & 0.83        \\
		\hline
		48.9 & 0.43 & 0.71 & 0.79 & 0.79        \\
		\hline

		\multicolumn{5}{l}{GENOMIC} \\
		\hline
		EMD & FedAvg & FedProx & CFL & FedGMCC \\
		\hline
		6.2 (real) & 0.88 & 0.89 & 0.91 & 0.93\\
		\hline
		10.5       & 0.88 & 0.91 & 0.93$\mid$ 0.92 & 0.93$\mid$0.93 \\
		\hline
		20.6       & 0.56 & 0.70 & 0.80$\mid$0.79 & 0.84$\mid$0.83 \\
		\hline
		26.3       & 0.21 & 0.65 & 0.80$\mid$0.79 & 0.84$\mid$0.83 \\
		\hline
	\end{tabular}

\end{table}

\section{Conclusion}
\label{sec:conclusions}

We presented \pname, a federated learning framework that consists of novel clustering rules and a new aggregation procedure. Substituting real datasets by a surrogate Monte Carlo dataset, we show how the curve finding procure can reveal the geometric connection between congruent models serving as a clustering rule of client model weights. Furthermore, \pname\ aggregation rule averages all the intermediary model weights on the curve parametrization, leading to better generalization 
and extending the classical FedAvg aggregation rule to weight spaces of curves training loss manifold. 

\pname\ outperformed other FL training algorithms on the EMNSIT62 and genomic sequence sensitivity classification tasks where we controlled the non-IID-ness using an artificial partitioning technique and the EMD measure. In high non-IID setting \pname\ yielded convergence rates respectively 36\% and 8\% faster than those of FedAvg and FedProx on EMNSIT62. \pname\ outperformed FedAvg, FedProx and CFL on the genomic datasets by 63$\%$, 19$\%$ and 4 $\%$.   

\printbibliography

\appendix

\section{Earth Mover Distance}
\label{apx:emd}

\begin{table*}[t!]
	\caption{Parameters and network configuration for autoencoders to accelerate EMD computation. Training progressed at a rate of $0.01$ for EMNIST62 and with $0.001$ for the different genomic datasets. We trained in batches of size $128$.}
    \label{table:autoencoder}
\footnotesize
	\centering
	\begin{tabular}{c|c|c|c|c|c}
		& Encoder Layers & dim(features) & Decoder layers & Learning rate & Reconstruction loss \\
		\hline
		EMNIST62 & 4 x CNN  & 7x7x2 & 4 x TransCNN & 0.01 & $\leq 0.0001 \%$ \\
		\hline
		SENSG-* & 2 x CNN + 2 x LSTM & 30x1 & 2 x LSTM + 2xTransCNN & 0.001 & $\leq 0.0001 \%$ \\
	\end{tabular}
\end{table*}

Earth Mover Distance (EMD) is defined as a distance metric between two distributions $\mathcal{A}$ and $\mathcal{B}$. It computes the minimal distance for mapping all clusters of distribution $\mathcal{A}$ (set of suppliers) to any cluster of distribution $\mathcal{B}$ (set of consumers). The distributions are thereby characterized by signatures $\mathcal{A} = \{ (p_1, w_{p_1}),...,(p_m, w_{p_m}) \}$ and likewise $\mathcal{B} = \{ (q_1, w_{q_1}),...,(q_n, w_{q_n}) \}$, where clusters are represented as bin centroids $p_i$ with weight $w_{p_i}$ (and $q_j$ with weight $w_{q_i}$, respectively). The overall cost for transferring all clusters from $\mathcal{A}$ to $\mathcal{B}$ is:

\begin{equation}
	C = \sum_{i\in\mathcal{A}}\sum_{j\in\mathcal{B}}c_{ij}f_{ij}
\end{equation}

where $c_{ij}$ is the ground distance between the supports $p_i$ and
$q_j$ and $f_{ij}$ is the flow between $p_i$ and $q_j$ that needs to
be minimized under the constraints:
\begin{enumerate}
\item $f_{ij} \geq  0$ (unidirectional flow),
\item $\sum_{(p_i, w_i) \in \mathcal{A}} f_{ij} \leq q_j$ (limited consumer storage),
\item $\sum_{(q_j, w_j) \in \mathcal{B}} f_{ij} \leq p_i$ (limited supply), and
\item $\sum_{(p_i, w_i) \in \mathcal{A}}\ \sum_{(q_j, w_j) \in \mathcal{B}} f_{ij} =
       \sum_{(p_i, w_i) \in \mathcal{A}} w_i =
       \sum_{(q_j, w_j) \in \mathcal{B}} w_j$ (transfer all).
\end{enumerate}

With optimal flow $f^{*} = \arg \min_f (C)$, EMD for a pair of distributions follows as:
\begin{equation}\label{eq:EMD}
	d(\mathcal{A},\mathcal{B})= \frac{\sum_{i\in\mathcal{A}}\sum_{j\in\mathcal{B}}c_{ij}f_{ij}^{*}}{\sum_{i\in\mathcal{A}}\sum_{j\in\mathcal{B}}f_{ij}^{*}}
\end{equation}

We extend $d(\mathcal{A},\mathcal{B})$ to a set of $K$ distributions $\{ \mathcal{D}_1,...,\mathcal{D}_K \}$ by averaging the set of pairwise EMDs between the $k$-th distribution $\mathcal{D}_k$ and the distribution over the whole population $\mathcal{D}$ $\mathcal{D}_k$, using

\begin{equation}
	d(\mathcal{D}_1,...,\mathcal{D}_K) =  \frac{1}{N} \sum_{k=1}^K  n_k d(\mathcal{D}_k,\mathcal{D}) 
\end{equation}
where $n_k$ is the sample size of the dataset generated by distribution $\mathcal{D}_k$, $N$ is the total population size and $\mathcal{D} = \frac{1}{N}\sum_k^K n_k \mathcal{D}_k$. In the context where $\mathcal{D}_k$ models a labeled dataset with $C$ classes, the joint probability distribution is factorized in terms of its conditional and marginal probability distribution $\mathcal{D}_k = p_k(\X,\y) = \{ p_k(y=i,\X) \cdot p_k(\X) \}$ for $i=1,...,C$, i.e, the EMD is computed over all the classes for a given feature $\X$.

To accelerate EMD computation we reduce datasets to their essentials by training autoencoders using a SGD optimizer and mean square error as reconstruction loss function. Table~\ref{table:autoencoder} shows the parameters for the autoencoders for the two benchmarks we used to evaluate our approach.

\section{Datasets}
\label{app:datasets}

\subsection{Image Datasets}

EMNIST62~\footnote{\url{https://colab.research.google.com/drive/1r-c6UTkJEQx3Pi-Hl9q_MoveIF_0h03M}}
is an image dataset for simulating non-IID image
classification \cite{cohen_afshar_tapson_schaik_2017}. It comprises a
set of 814255 handwritten alpha-numeric characters, labeled as 62
unbalanced classes, formatted in 28x28 pixel images.

\subsection{Genomic Datasets}

We used two different genomic datasets: SENSG-R and SENSG-S. \\
SENSG-R is composed of 7 500 000 reads from four randomly selected genomes from each of the three major populations represented in the 1000 Genomes Project (1000GP)~\footnote{\url{https://www.internationalgenome.org/}}: African, European and Asian(see Table~\ref{tab:superpop_genomes}). With this we intend to resemble the natural representation one would obtain when sampling in regions where these populations are dominant.\\
\textbf{SENSG-S} is composed of one million reads generated from twenty individual genomes ($10^5$ reads for each genome) and the randomly selected individuals are the following: HG00096, HG00097, HG00099, HG00100, HG00101, HG00102, HG00103, HG00105, HG00106, HG00107, NA21128, NA21129, NA21130, NA21133, NA21135, NA21137, NA21141, NA21142, NA21143, and NA21144.

\begin{table}[h!]
	\caption{Genomes used per population dataset.}
    \label{tab:superpop_genomes}
\footnotesize
	\centering
	\begin{tabular}{c|c|c}
	ASIA & AFRI & EURO \\
	\hline
	HG00543 & HG02703 & HG00315 \\
	\hline
	HG00559 & HG02769 & HG00327 \\
	\hline
	HG00566 & HG02715 & HG00334 \\
	\hline
	HG00578 & HG02771 & HG00339 \\
	\hline
	HG00581 & HG02722 & HG00341 \\
	\hline
	HG00580 & HG02676 & HG00346 \\
	\hline
	HG00593 & HG02808 & HG00353 \\
	\hline
	HG00598 & HG02810 & HG00358 \\
	\hline
	HG00592 & HG02614 & HG00360 \\
	\hline
	HG00613 & HG02839 & HG00365 \\
	\end{tabular}
\end{table}

For generating these datasets, we follow three steps: \\
\textbf{Step 1 - Genomic sequence generation: } We compiled the chromosome 1 sequence of each individual selected by combining the human reference genome GRCh37 and the individual's variants in the 1000 GP. Second, we randomly select positions in the sequence to use as seed for obtaining 150 character sequences (reads), comprised of the nucleotides A, T, G, and C. 
Reads constitute the first digitized information obtained from next
generation sequencing machines, which are subsequently processed to
extract variations (i.e., what distinguishes us one from another and
what might carry sensitive information, like, disease prepositions). 
Therefore, genomic data is a practical example where data is generated in multiple geo-distributed locations, e.g., hospitals in different continents, and it must not be shared among different locations for privacy reasons.\\
\textbf{Step 2 -- Labeling:} We labeled each nucleotide by deeming it as sensitive if a genomic variation reported in the 1000 GP or insensitive otherwise, respectively, 1 or 0. 
For SENSG-S only, we simulate a concept shift by flipping some labels manually in order to obtain non-IID-ness. \\
\textbf{Step 3 -- Reads encoding:} Next, we used word2vect encoding to convert the genomic sequence in a vector. After, this step the reads are ready to be used for the training. 
In this step we consider a vocabulary (word size) of 5 letters and we generated all the consecutive 5 nucleotides sequences of each read. 

\subsection{Artificial Partitioning} 
SENSG-A and EMNIST62 are partitioned artificially into $K$ subsets
using our iterative clustering and distribution algorithm, which we
describe in the following. We first apply $k$-mean clustering to
divide the dataset into $K$ subsets with maximal EMD. Then, computing
the center of gravity (COG) of all subsets, we randomly select
elements from the farthest subset $D_i$ and assign them to other
subsets $D_j$. This reduces the inter-subset distance $d(D_i,
D_j)$  and hence the overall EMD. We retain the current partition if pairwise EMD values ($d(D_i, D_j)$) are normally
distributed (according to the Shapiro normality test). We repeat this process until a we found a partition with a total EMD value that is low enough. 

\end{document}